\documentclass[10pt, logo, twocolumn, copyright]{nvidiatechreport}
\usepackage{wasysym}
\usepackage{threeparttable}

\renewcommand{\today}{}

\newcommand{\yes}{\ding{51}}
\newcommand{\no}{-}
\newcommand{\prt}{$\sim$}
\newcommand{\hum}{\CIRCLE}
\newcommand{\mix}{\LEFTcircle}
\newcommand{\aut}{\Circle}
\newcommand{\rot}[1]{\rotatebox{90}{#1}}
\newcommand{\fn}[1]{\makebox[0pt][l]{$^{#1}$}}

\newcommand{\supp}{supplementary material}
\usepackage[numbers,sort&compress]{natbib}
\titlespacing*{\paragraph}{0pt}{0.5ex plus 0.0ex minus 0.5ex}{1em}
\titleformat{\subsubsection}
{\bfseries}
{\thesubsubsection.}
{0.5em}
{#1}
[]

\title{CASCADE: A Spatio-Temporal-Causal Reasoning Representation and Dataset for Driving}
\author[1]{Jenny Schmalfuss}
\author[1]{Despoina Paschalidou}
\author[2,*]{Simon Gerstenecker}
\author[1]{German Ros}
\author[1]{Jose M. Alvarez}
\affil[1]{NVIDIA}
\affil[2]{ETH Zürich}
\affil[*]{Work done during NVIDIA Internship}

\begin{abstract}
Reasoning is a promising route to the generalization that autonomous driving requires in the long tail, as it can infer how the elements of a scene depend on one another and traverse those dependencies to conclusions beyond what is observed.
Yet it is hard to tell whether a model's conclusions follow the scene's dependencies, because no driving representation makes them explicit enough to test against.
Text-based reasoning traces lack spatio-temporal grounding, spatio-temporal scene graphs lack causal links, and reasoning annotations at scale are increasingly model-generated and hard to verify.
To this end, we introduce CASCADE (Causal Spatio-Temporal Analysis of Driving Environments), which encompasses two components: (1) a structured scene representation for reasoning in driving scenes and (2) a human-annotated dataset built on it.
For every actor that interacts with the ego vehicle, the CASCADE representation records frame-by-frame, for as long as the actor is visible, what action is taken, where it occurs, and how it depends on the actions and states of others.
The resulting structure makes reasoning predictions machine-verifiable: they can be scored against it element by element, without relying on (M)LLM judges.
The CASCADE dataset provides comprehensive human annotations for 2,066 driving clips of the PhysicalAI dataset, with over 34K elements that establish the spatio-temporal and causal context of each scene, including 8.6K time-stamped ego and agent actions, 3.7K causal links and 2.9K potential influences, and 6.1K annotations for agents, objects, traffic lights, and environments.
Being entirely human-annotated, CASCADE provides the reference for this comparison: benchmarking the reasoning abilities of Physical AI models, and verifying the quality of automatically generated reasoning labels.
The CASCADE dataset is available at \url{https://huggingface.co/datasets/nvidia/cascade}.
\end{abstract}

\begin{document}
\maketitle

\section{Introduction}
\label{sec:intro}

Recently, physical AI models for driving have matured in perception capability and model architecture.
But to close the gap to L4 driving \cite{sae_j3016_2021}, they must handle rare situations, which training data covers too sparsely for reliable generalization \cite{scholkopf2021causal}.
\textbf{Reasoning} offers a promising avenue: a reasoning model combines its perception of scene elements with knowledge of how they depend on one another \cite{bottou2014machine}, and traverses these connections to situation-adaptive conclusions \cite{huang2023reasoning}.
\begin{figure}
\centering
  \includegraphics[width=.7\columnwidth]{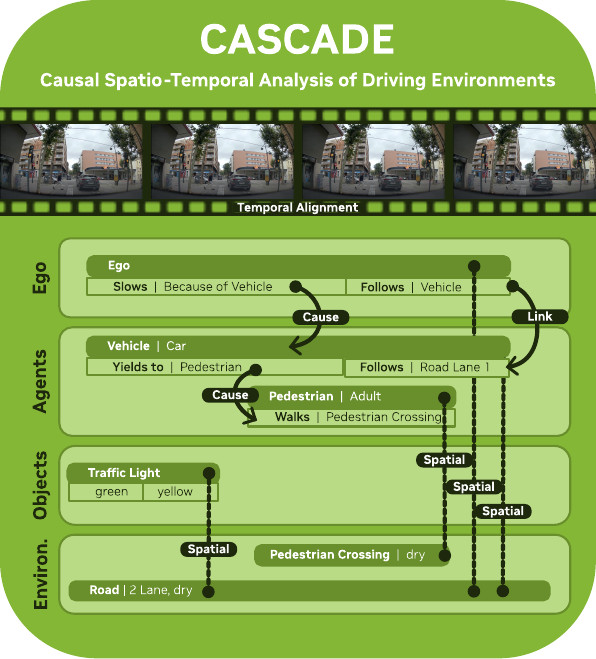}%
    \caption{\textbf{CASCADE reasoning scene graph} for driving scenes. It captures actors, like ego vehicle and agents, together with their actions and properties for the full duration of the clip. Start and end times for visibilities and actions enable temporal grounding. For spatial grounding, actors and objects are anchored against environment annotations.
  Causality is established via cause-effect arrows from the caused action to its cause.}
  \label{fig:teaser}%
\end{figure}
In driving, reasoning is assessed via text reasoning traces, Chain-of-Thought (CoT) or Chain-of-Causation (CoC) \cite{nie2024reason2drive, kim2018textual, wang2025alpamayo}, and increasingly used as supervision.
But to assess how well models recover which scene elements an action depends on, two prerequisites are missing.
First, no scene representation makes reasoning elements explicit and verifiable: text traces \cite{nie2024reason2drive, kim2018textual, wang2025alpamayo} lack grounding in space and time \cite{zhang2024multimodal, xiao2024trust}, and cannot be verified against ground truth, while spatio-temporal scene graphs (STSGs) \cite{ji2020actiongenome, singh2022road, mlodzian2023nuscenes, malawade2022roadscene2vec} are grounded but record co-occurrence rather than why an action happened.
Second, no annotation combines complete reasoning context with human provenance: human attribution datasets~\cite{ramanishka2018honda, you2020traffic} cover only direct causes, while reasoning supervision at scale is model-generated \cite{wang2025alpamayo} and frequently scene-contradicting \cite{mayumu2026faithful}, risking a capability-degrading recursion \cite{shumailov2024ai}.
Assessing reasoning progress thus hinges on a comprehensively grounded \textbf{reasoning scene representation} to verify model predictions, and a context-complete, human-annotated \textbf{ground truth dataset} to evaluate reasoning models and the auto-annotations they are trained on.

Prior reasoning representations and datasets fall short on four requirements.
One, the scene representation should be \emph{spatio-temporally-causally grounded} to anchor where, when, and why an action happened, and ensure that reasoning steps are verifiable against the scene.
Text-based reasoning traces lack spatial and temporal grounding, which is associated with hallucinated or ungrounded rationales \cite{zhang2024multimodal, xiao2024trust}.
STSGs are spatio-temporally grounded but lack causal connections, and thus descriptive instead of explanatory \cite{ji2020actiongenome, singh2022road, mlodzian2023nuscenes}.
Two, the scene representation should render model outputs \emph{machine-verifiable}, \ie scoreable against ground truth.
Existing free-form reasoning traces cannot be matched and are scored by (M)LLM judges \cite{wang2025alpamayo}, inheriting their biases and inconsistencies \cite{ye2025justice, shi2025judging,vs2026ghost}.
While structured benchmarks circumvent the issue in other domains \cite{white2025livebench,schmalfuss2025parc}, reasoning for driving currently favors free-form text.
Three, the dataset should capture the ego vehicle's \emph{complete causal context}, \ie include (i) actors and objects with indirect ego-action connections (full causal chain), and (ii) their full action and property traces over time (temporal completeness).
For reasoning, this is critical: The ego should slow down in anticipation of a pedestrian approaching a crossing, as opposed to emergency braking if it appeared from an occlusion.
Yet prior causal data only records triggering actions \cite{wang2025alpamayo, ramanishka2018honda, malla2023drama}, and action-traced data \cite{singh2022road, sachdeva2024rank2tell} omits causes.
Four, the dataset should provide \emph{human annotations}.
Existing reasoning annotations are often (M)LLM-generated for scale \cite{wang2025alpamayo}.
But (M)LLMs misinterpret scenes \cite{wen2023road}, are frequently unfaithful \cite{turpin2023language}, and no match for expert assessment \cite{huang2026nureasoning}.
Pure automation misses the required structure, and even map-derived scene graphs rely on humans to select relevant elements \cite{wild2026bridging}.
Though indispensable for reasoning benchmarking, human ground truth is scarce due to acquisition cost.

\begin{figure*}
    \centering
    \includegraphics[width=1\linewidth]{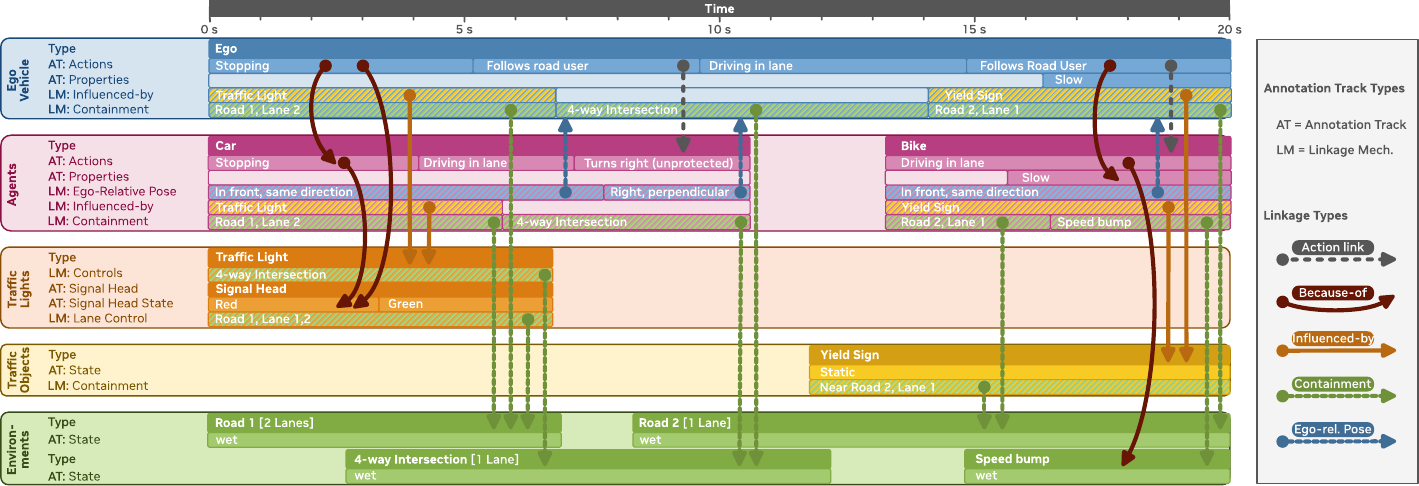}
    \caption{\textbf{CASCADE annotated reasoning scene graph.} Scenes are hierarchical spatio-temporal-causal scene graphs. On the broadest level, they are composed of five scene entity types: the ego vehicle, agents, traffic lights, traffic objects, and environments, which have static attributes like types or quantities. On a finer level, these entities come with temporal annotation tracks to describe their actions, properties and states, as long as they are active in the clip. On the same level, entities of different types can be connected to one another via linkage mechanisms. Linkage mechanisms are temporal linking tracks, represented as striped annotations in the colors of the linked entity types, and direct arrows, connecting two annotation tracks.}
    \label{fig:CASCADE_schema_doublecol}
\end{figure*}

To address these four gaps, we introduce \textbf{CASCADE} (Causal Spatio-Temporal Analysis of Driving Environments).
CASCADE supplies both principal components: a \textbf{spatio-temporal-causal scene graph} representation and a corresponding \textbf{human-annotated dataset}.
Designed for reasoning evaluations in driving, they capture the whole driving scene with its decision-critical entities, their actions, interactions, and locations frame-by-frame, \cf \cref{fig:teaser,fig:CASCADE_schema_doublecol}.

\textbf{CASCADE's reasoning scene graph} represents a driving scene as typed nodes connected by text-descriptive edges.
Nodes cover five entity types -- the ego vehicle, agents, traffic lights, traffic objects, and environments -- each with its actions, properties, and states while visible.
Edges express causes (because-of, influenced-by) and spatial relations (containment, ego-relative pose).
This structure \emph{grounds reasoning} spatially, temporally, and causally: spatially through environments, ego-relative poses, and image track points; temporally through time segments bounding all actions, properties, and states; and causally through because-of and influenced-by edges.
The causal dimension distinguishes CASCADE from STSGs \cite{ji2020actiongenome, singh2022road, mlodzian2023nuscenes}, which are spatio-temporally descriptive but lack explanations.
With typed and anchored elements, the graphs are \emph{programmatically verifiable} against ground truth element by element, as in causal discovery \cite{chen2024mecd, chen2026mecdplus}.

\textbf{CASCADE's human-annotated dataset} instantiates this reasoning scene graph representation for complex real-world driving clips.
Its annotations capture the ego's \emph{complete causal context}, differing from prior annotations in two ways.
Full causal chains record all causally-linked actors and objects in a bi-directional manner, \ie \texttt{ego stops} $\!\leftarrow\!$ \texttt{van stops} $\!\leftarrow\!$ \texttt{stop sign} if ego stops for a van at a stop sign, and \texttt{ego drives} $\!\rightarrow\!$ \texttt{car yields} if ego actions affect other agents downstream.
Temporal completeness then records the full action and property traces for those agents and objects.
All CASCADE annotations are created and reviewed by \emph{human experts}, no part is model-annotated.
At 0.5-2h of expert effort per 20s clip, such dense human-adjudicated ground truth is rare: CASCADE holds over 34,000 reasoning-relevant elements across 2,066 clips of the PhysicalAI dataset \cite{nvidia2025physicalai}.
At a time when reasoning data is largely auto-generated for scale, CASCADE is a reference that model predictions and generated labels alike can be scored against without a judge, paving the way for reliable reasoning supervision.

\paragraph{Contributions} of CASCADE are two-fold:
\vspace*{-.5\baselineskip}
\begin{enumerate}
    \item \textbf{CASCADE's scene representation}, a frame-by-frame reasoning scene graph for driving with spatio-temporal-causal grounding, against which predictions are verifiable without judges.
    \item \textbf{CASCADE's dataset} of 2,066 human reasoning annotations, capturing complete causal context of all actors and objects on the ego's causal chain, with their actions and properties over time.
\end{enumerate}
\vspace*{-.5\baselineskip}
We further analyze the dataset's action and property distributions, causal chain lengths, and temporal completeness.

\section{Related Work}
\label{sec:related_work}

\begin{table}[t]
\centering
\small
\setlength{\tabcolsep}{4pt}
\caption{\textbf{Driving-video reasoning representations and datasets,}
following the four requirements grounding, verifiability (Verif.), complete causal context (Caus.\ con.), and human annotation (Hum.).
All rows are ego-centric video datasets.
Symbols: S spatial, T temporal, C causal, FbF frame-by-frame, MH multi-hop, BD bidirectional, AT action traces, \yes~yes, \prt~partly, \no~no, \hum~human, \mix~mix, \aut~automatic.}
\label{tab:related}
\centering
\begin{threeparttable}
\scalebox{0.82}{%
\begin{tabular}{@{}cl c@{\hspace{3pt}}c@{\hspace{3pt}}c@{\hspace{3pt}}c c c@{\hspace{1.5pt}}c@{\hspace{1.5pt}}c c@{}}
\toprule
& & \multicolumn{5}{l}{\textbf{Representation}} & \multicolumn{4}{l}{\textbf{Dataset}} \\
\cmidrule(lr){3-7}\cmidrule(lr){8-11}
& & \multicolumn{4}{c}{\textbf{Grounding}} & \textbf{Verif.} & \multicolumn{3}{c}{\textbf{Caus.\ con.}} & \textbf{Hum.} \\
\cmidrule(lr){3-6}\cmidrule(lr){7-7}\cmidrule(lr){8-10}\cmidrule(lr){11-11}
&  & S & T & C & FbF & & MH & BD & AT & \\
\midrule
\multirow{5}{*}{\rot{\textbf{Text traces}}}
& Alpamayo-R1~\cite{wang2025alpamayo} & \no & \yes\!/\!\no\fn{c} & \yes & \no & \prt & \no & \no & \no & \aut\fn{a} \\
& nuReasoning~\cite{huang2026nureasoning} & \yes & \yes\!/\!\no\fn{c} & \yes & \no & \prt & \no & \no & \no & \mix \\
& Reason2Drive~\cite{nie2024reason2drive} & \yes & \no & \yes & \no & \prt & \no & \no & \no & \mix \\
& LingoQA~\cite{marcu2024lingoqa} & \no & \no & \yes & \no & \no & \no & \no & \no & \mix \\
& BDD-X~\cite{kim2018textual} & \no & \yes & \yes & \no & \no & \no & \no & \no & \hum \\
\midrule
\multirow{6}{*}{\rot{\textbf{Causal attrib.}}}
& SocioDrive~\cite{yan2026causaldrive} & \no & \no & \yes & \no & \prt & \no & \yes & \no & \aut \\
& BDD-OIA~\cite{xu2020bddoia} & \no & \no & \yes & \no & \yes & \no & \no & \no & \hum \\
& CTA~\cite{you2020traffic} & \no & \yes & \yes & \no & \yes & \no & \no & \no & \hum \\
& DRAMA~\cite{malla2023drama} & \yes & \no & \yes & \no & \prt & \yes\fn{b} & \no & \no & \hum \\
& HDD~\cite{ramanishka2018honda} & \yes & \yes & \yes & \no & \yes & \no & \no & \no & \hum \\
& Rank2Tell~\cite{sachdeva2024rank2tell} & \yes & \yes & \yes & \prt & \prt & \no & \no & \no & \hum \\
\midrule
\multirow{6}{*}{\rot{\textbf{Scene graphs}}}
& roadscene2vec~\cite{malawade2022roadscene2vec} & \yes & \yes & \no & \yes & \yes & \no & \no & \no & \aut \\
& OpenLane-V2~\cite{wang2023openlanev2} & \yes & \yes & \no & \prt & \yes & \no & \no & \no & \mix \\
& nSKG~\cite{mlodzian2023nuscenes} & \yes & \yes & \no & \prt & \yes & \no & \no & \no & \mix \\
& CRS~\cite{wild2026bridging} & \yes & \yes & \no & \prt & \yes & \no & \no & \no & \mix \\
& ROAD-Waymo~\cite{singh2024roadwaymo} & \yes & \yes & \no & \yes & \yes & \no & \no & \yes & \mix \\
& ROAD~\cite{singh2022road} & \yes & \yes & \no & \yes & \yes & \no & \no & \yes & \hum \\
\midrule
& \textbf{CASCADE} (ours) & \yes & \yes & \yes & \yes & \yes & \yes & \yes & \yes & \hum \\
\bottomrule
\end{tabular}
}
\begin{tablenotes}\footnotesize
\item[a] GPT-5-labelled; $\sim$10\,\% non-public human-labelled subset.
\item[b] Second hop is free text, depth $\leq$2.
\item[c] Actions/states and causes annotated at different densities.
\end{tablenotes}
\end{threeparttable}
\end{table}

We first motivate graphs as reasoning representations, then review prior driving datasets along the four requirements for representations: grounding, verifiability, causal context and human annotation, \cf \cref{tab:related}.

\paragraph{Structure in reasoning representations.}
There are three forms of reasoning representations for driving: \emph{causal attributions} \cite{ramanishka2018honda,xu2020bddoia,you2020traffic,malla2023drama,sachdeva2024rank2tell} state causes for ego-actions via labels or boxes, \emph{reasoning traces} \cite{kim2018textual,nie2024reason2drive,marcu2024lingoqa,wang2025alpamayo,huang2026nureasoning} give text explanations or chains of thought, and \emph{scene graphs} \cite{singh2022road,singh2024roadwaymo,mlodzian2023nuscenes,malawade2022roadscene2vec,wang2023openlanev2,wild2026bridging} have typed element-nodes with relations.
Driving models appear to move towards more structured reasoning along these three representations:
from closed-vocabulary explanation heads~\cite{xu2020bddoia}, over free-text justifications~\cite{xu2024drivegpt4} and chain-of-thought traces~\cite{tian2024drivevlm,wang2025alpamayo,ishaq2025drivelmmo1}, to graph-organized question chains~\cite{sima2024drivelm} and scene-graph conditioned models~\cite{schmidt2025graphpilot,zhang2025graphad}.
Outside driving, graphs established structure for visual reasoning.
Image scene graphs describe objects with attributes and their relations~\cite{johnson2015retrieval,krishna2017visual}, and are generators for QA with exact answers~\cite{hudson2019gqa} as well as reasoning substrate \cite{shi2019xnm,koner2021graphhopper}.
Video scene graphs extend this to time~\cite{shang2017vidvrd,shang2019vidor,ji2020actiongenome,yang2023pvsg,yang2023fourDpsg,rodin2024easg} and video QA generation~\cite{wu2021star,yu2023anetqa}, but their edges record what co-occurs, keeping those graphs descriptive.
Explanatory, causal graphs have matured for text~\cite{wang2022mavenere,ning2018joint}, and exist for synthetic~\cite{mao2022clevrerhumans} and real \cite{chen2024mecd,chen2026mecdplus} general video.
CASCADE introduces explanatory graphs for \emph{driving} videos.

\paragraph{Representation grounding.}
Prior driving representations ground reasoning spatially, temporally or causally.
Text reasoning \cite{nie2024reason2drive,kim2018textual,marcu2024lingoqa,wang2025alpamayo,huang2026nureasoning} causally grounds ego actions, but misses temporal grounding unless object references were in the source dataset, and lacks temporal grounding as traces cover entire clips or keyframes.
Causal attributions \cite{ramanishka2018honda,xu2020bddoia,you2020traffic,malla2023drama,sachdeva2024rank2tell} are causally and often spatially grounded through boxes on the causing element, but miss true temporal grounding with causes attached to clips or coarse segments.
Scene graphs of road users and their actions \cite{singh2022road,singh2024roadwaymo,mlodzian2023nuscenes,malawade2022roadscene2vec}, road topology~\cite{wang2023openlanev2,li2026toponet,wu2024topomlp,li2024lanesegnet,shin2025instagram}, or both~\cite{wild2026bridging} are spatially and (partly) frame-by-frame temporally grounded, but miss causal grounding in their relations.
Frame-by-frame temporal grounding thus exists without causal grounding, and causes only with coarse temporal grounding.
CASCADE is 3-fold grounded through causes anchored in space and frame-by-frame time.

\paragraph{Representation verifiability.}
Reasoning traces in driving \cite{wang2025alpamayo,ishaq2025drivelmmo1,marcu2024lingoqa,chen2025codalm,tian2024drivevlm,song2026drivecritic,xiong2026phycritic} are scored by (M)LLM judges \cite{zheng2023judging}. 
But their judgments exhibit systematic biases \cite{ye2025justice,shi2025judging,tan2025judgebench,li2026preferenceleakage}, are unstable~\cite{vs2026ghost}, and zero-shot VLMs cannot judge driving behavior as well as humans~\cite{sun2026drivejudge}.
Several recent benchmarks therefore avoid judges if the answer format permits~\cite{white2025livebench,mialon2024gaia,schmalfuss2025parc,huang2026nureasoning}.
Graph structured ground truth can be scored deterministically: topology by edge prediction~\cite{wang2023openlanev2}, and causal graphs by edge accuracy~\cite{chen2024mecd}.
CASCADE's graph reasoning representation is thus machine-verifiable against ground truth element by element.

\paragraph{Dataset complete causal context.}
Driving video datasets generally capture first-degree causes for ego vehicle actions \cite{ramanishka2018honda,xu2020bddoia,sachdeva2024rank2tell,you2020traffic,wang2025alpamayo,huang2026nureasoning}, only DRAMA \cite{malla2023drama} adds a second-level textual cause.
The ego's effect on others is only recorded by two works: SocioDrive-Bench~\cite{yan2026causaldrive} for video, WOMD-Reasoning~\cite{li2025womdreasoning} for trajectory data, but both are auto-annotated.
Conversely, datasets that trace actions of all road users over time~\cite{singh2022road,singh2024roadwaymo} do not capture causes.
In contrast, CASCADE records the complete causal context: causal chains of any depth in both directions around the ego, together their actor's full action traces.

\paragraph{Dataset human annotation.}
Among driving data annotations, human annotations are associated with the highest label quality~\cite{liu2024addatasets}.
Annotations span fully automatic~\cite{wang2025alpamayo,yan2026causaldrive,arai2025covla,malawade2022roadscene2vec}, mixtures of auto-generation and human verification \cite{huang2026nureasoning,ishaq2025drivelmmo1,chen2025codalm,kim2025vruaccident,fruhwirth2025stsbench}, rule derivation from human ground truth \cite{li2025womdreasoning,qian2024nuscenesqa,park2025nuplanqa,mlodzian2023nuscenes,nie2024reason2drive,fruhwirth2025stsbench,sima2024drivelm}, or human labels on inherited elements \cite{singh2024roadwaymo,wang2023openlanev2,wild2026bridging},  and fully human annotation~\cite{ramanishka2018honda,xu2020bddoia,you2020traffic,malla2023drama,sachdeva2024rank2tell,kim2018textual,singh2022road}.
Among those, CASCADE is the first driving dataset with fully human-annotated causal context.

\cref{tab:related} summarizes reasoning representations and datasets for driving video, which highlights each representation family's failure modes:
Text traces and causal attributions are 1st-degree causally grounded, but rarely spati-temporally grounded and verifiable, and lack full action traces.
Spatio-temporal scene graphs are grounded and verifiable, but lack causality.
Importantly, each family can be derived from CASCADE: its representation is a scene graph, its because-of chains read as reasoning traces, and their first links are causal attributions.

\section{CASCADE}

In the following, we introduce CASCADE with its two components: Its reasoning scene representation and its human-labelled dataset.
We first describe the CASCADE's reasoning scene graph to represent the spatio-temporal-causal dependencies of actors within a driving scenario. 
Then, we detail the dataset collection process and give an overview of the CASCADE dataset and its statistics.

\subsection{CASCADE's Reasoning Scene Graph}

\begin{figure}[tb]
    \centering
    \includegraphics[width=\linewidth]{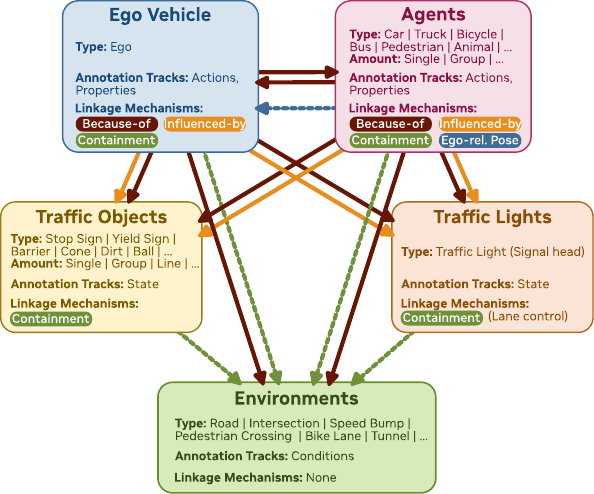}
    \caption{\textbf{CASCADE scene graph schema.} The schema is based on five entity types (Ego, Agents, Traffic Objects, Traffic Lights and Environments) that come with fixed properties (Types and Amounts), annotation tracks to capture actions and properties over time as long as they are active, and linkage mechanisms to establish spatial and causal connections between actions, properties or environments. Causal linkage mechanisms are because-of and influenced-by links, while spatial linkage mechanisms are containment and ego-relative poses. Also see \cref{fig:CASCADE_schema_doublecol} for an annotation example using the schema.}
    \label{fig:CASCADE-annotation-schema}
\end{figure}

CASCADE represents each driving clip as a hierarchical spatio-temporal-causal scene graph.
A schematic overview of such a scene graph is in \cref{fig:CASCADE_schema_doublecol}, the components of the  world representation with all their possible interactions are in \cref{fig:CASCADE-annotation-schema}.

\paragraph{High-level representation structure.}
At the top level, the representation contains scene \emph{entities}, including the ego vehicle, other agents, traffic objects, traffic lights, and environments.
These entities carry stable descriptors, such as their type, quantity, lane count, or directionality.
Additionally, entities may have a varying number of temporal \emph{annotation tracks} that describe how their behavior or state changes over the clip, and \emph{linkage mechanisms} that may be temporal tracks or directed arrows and describe how spatial contexts, or influences change; \cf \cref{fig:CASCADE_schema_doublecol,fig:CASCADE-annotation-schema}.

All temporal annotations in CASCADE are represented as via segments in so-called tracks, where segments have start and end timestamps, \cf \cref{fig:CASCADE_schema_doublecol}.
We use the term \emph{dense} to describe tracks that are required to cover their relevant interval, \ie the period where the entity is visible, without gaps.
Examples are the ego and agent action tracks or traffic-light states.
Other tracks are \emph{sparse}: they are instantiated only when the corresponding attribute is present, \eg ego being slow, or actors being influenced by a sign.

CASCADE derives its causal and spatial structure from linkage mechanisms between entity records and their temporal segments.
These relations connect otherwise parallel entity timelines into an interconnected spatio-temporal-causal graph, for example by linking a stop action to a red traffic light or a yield action to the agent it yields to.

\begin{figure}
    \centering
    \includegraphics[width=\linewidth]{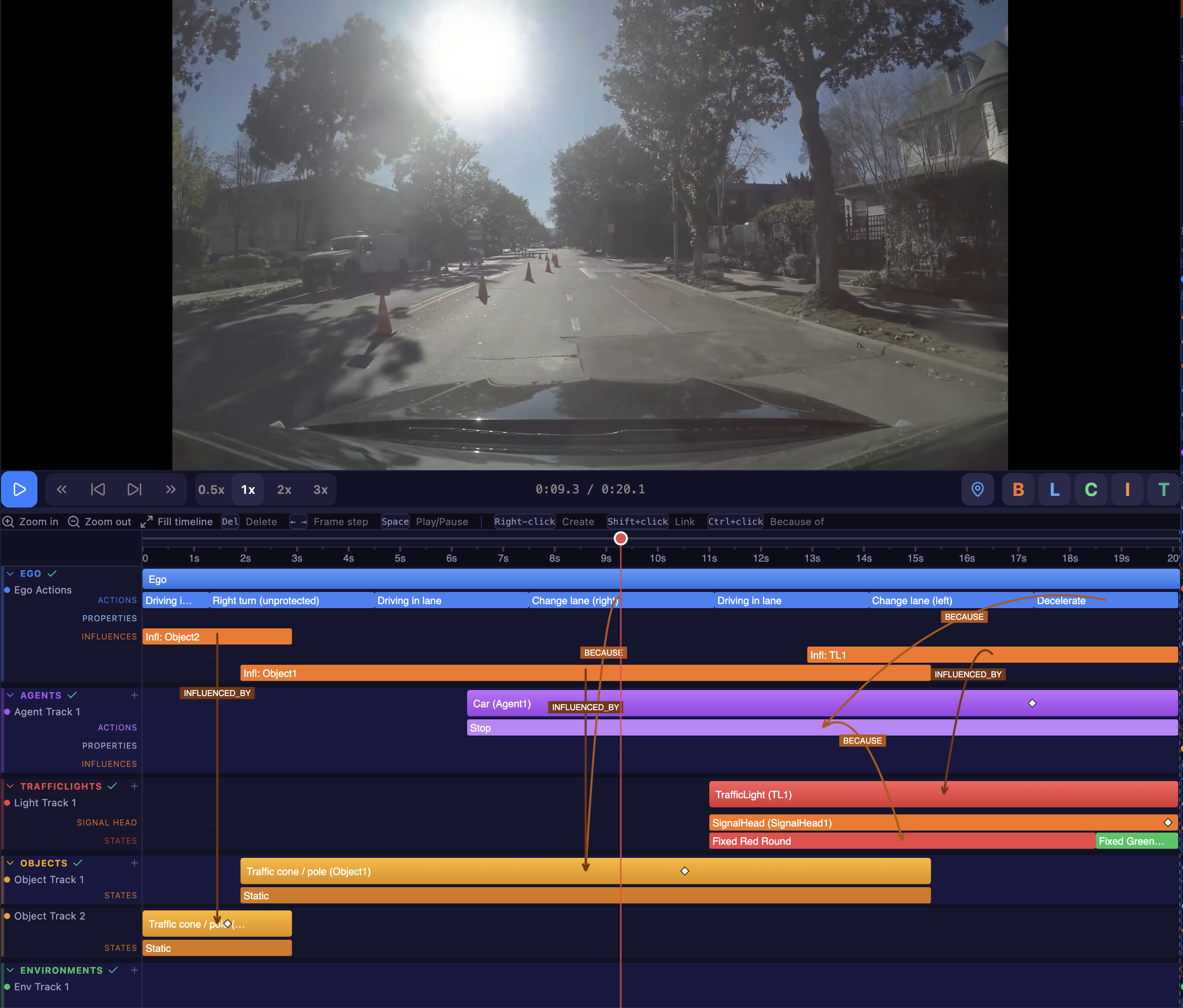}
    \caption{\textbf{CASCADE annotation example.} At the displayed time, shown as red vertical line on timeline, the ego vehicle changes lanes because of the traffic cone line, before it decelerates for a stopped car at a red light.}
    \label{fig:CASCADE-sample-annotation}
\end{figure}

The example in \cref{fig:CASCADE-sample-annotation} illustrates these concepts for a real CASCADE annotation, and shows how the ego vehicle changes lanes to the right because of traffic cones in its path.
In the following, we explain the CASCADE scene representation structure with its entity types, annotation tracks and linkage mechanisms.
An exhaustive description of all representation fields and options in the \supp, \cref{sec-supp:schema}.

\subsubsection{Representation Entities}
CASCADE represents the main components of a driving scene as five classes of top-level entities.
These entities are described via individual collections of persistent attributes, temporal annotation tracks, and relational linkage mechanisms, \cf \cref{fig:CASCADE-annotation-schema}.
Entities are annotated according to their visibility in the clip and include keypoints for visual grounding.

\paragraph{The ego vehicle} is the actively driving vehicle.
It has attributes beyond its type (ego vehicle), but owns rich annotation tracks like action and properties, and linkage mechanisms like containment, influenced-by and because-of links.

\paragraph{Agents} represent external actors that interact with or influence the ego.
Each agent has an intrinsic type, such as vehicle, pedestrian, animal, or other, and an amount descriptor distinguishing a single actor from a row or a group.
Agents own annotation tracks for actions and properties, and linkage mechanisms for containment, ego-relative pose,  spatial, influenced-by and because-of links.

\paragraph{Traffic lights} separate the traffic-light structure from the signals it displays.
The top-level traffic-light entity links to the environment controlled by the full system, while nested \textit{SignalHeads} represent distinct traffic signals.
The signal heads come with individual annotation tracks for their corresponding light states, and linkage mechanisms to controlled lanes.
For example, one Signal Head can show a green arrow for a left-turn lane while another shows a red light for the remaining lanes.

\paragraph{Traffic objects} represent static, regulatory, or portable scene elements that can affect driving behavior, such as signs, cones, barriers, or debris.
Some object types carry a quantity descriptor, such as single, group, line/row, channelizing line, or perimeter.
Objects own annotation tracks for their states and linkage mechanisms for containment.

\paragraph{Environments} represent road-infrastructure regions such as roads, intersections, roundabouts, tunnels, sidewalks, and cycle lanes.
Their stable descriptors include lane count and directionality where applicable, and they have annotation tracks for conditions.
They do not own explicit linkage mechansims, but instead act as spatial anchors for the containment relations of all other entity classes.

\subsubsection{Representation Annotation Tracks}
To extend beyond the static entity attributes, temporal annotations describe how entities act, change, or move in the environment over time.
CASCADE uses temporal \emph{annotation tracks} for behavior, actor properties, signal and object states, and environment conditions.

\paragraph{Actions} are the primary behavioral annotations for the ego vehicle and agents.
They are annotated densely over the relevant interval, so that each actor's behavior is continuously described while it is present in the scene.
Action types cover maneuvers such as stopping, yielding, following, lane changes, turns, nudging, overtaking, parking, walking, or running, depending on the actor type.
Actions can be marked with an illegal flag when they violate traffic rules or expected road behavior.
They are also the main entry point for causal annotation: they can start \emph{because-of} links to scene entities or other actions that led to the respective behavior.

\paragraph{Actor properties} describe temporary attributes of the ego vehicle or agents that may overlap with actions, such as speed, driving style, status, or signaling.
Signaling properties capture inter-actor communication through flashing lights, hand gestures, or held signs, and can optionally identify the actor to whom the signal is directed.

\paragraph{Traffic-light states} are densely annotated for each \textit{SignalHead}.
They describe the displayed signal over time, including activation type, color, and shape.

\paragraph{Object states} describe whether a traffic object is static or moving, with optional open/closed states for object types where this applies.

\paragraph{Environment conditions} describe temporary or local properties of environment entities, such as construction zones, wet or snowy surfaces, overgrown areas, or obscured lane markings.
If a condition does not apply for the duration, its timestamps follow the interval during which the ego vehicle is located in the affected environment, e.g., while it drives through a construction zone.

\subsubsection{Representation Linkage Mechanisms}
The central goal of CASCADE is to connect parallel entity timelines into a causal spatio-temporal graph.
To do so, CASCADE annotates several kinds of relations between entities and temporal annotations.
These relations capture direct causal interactions, broader influences on actor behavior, action targets, and spatial grounding in the road layout.

\paragraph{Because-of links} relate an action to its direct causes.
These causes can be other actions, actor properties, traffic-light states, traffic objects, environments, or annotator-defined causes.
By allowing actions to reference other actions as causes, because-of links can form causal chains across multiple entities to describe complex dynamic dependencies.

\paragraph{Influenced-by links} link the ego vehicle or agents to traffic lights or traffic objects that affect their decision-making without necessarily causing a specific action.
Examples include a green traffic light, yield sign, line of traffic cones, or construction equipment.
Influences are represented as sparsely annotated temporal segments with their own start and end timestamps, and multiple influences can apply to the same actor at the same time.

\paragraph{Action and signaling targets}
identify the entity that an annotation is directed toward without making it a causal explanation.
For example, following, overtaking, nudging, entering, and exiting actions can reference their target road user, object, or ego vehicle, while signaling properties can identify the actor to whom a gesture, flashing light, or held sign is addressed.

\paragraph{Spatial containment}
is the main spatial relation linking all entities to the environments and lanes they occupy. Spatial containments are represented using densely annotated time segments, each specifying the environment and lane occupied during that interval. There can be multiple containment relations at a time, representing that an actor is contained in multiple environments at once (e.g. cycle lane and road, pedestrian crossing and intersection). Additionally, spatial containment relations have the optional modifiers \textit{Near} and \textit{Left/Right edge}, which captures relevant spatial links beyond strict containments, such as a pedestrian standing near a crosswalk, parked cars on the right edge of a unmarked road or a traffic sign being near the intersection it controls. Containments can also be marked "Illegal" to represent entities occupying environments where they should not be located.
For traffic lights, the containment relation specifies the environment controlled by the traffic light itself and the lanes controlled by individual \textit{SignalHeads}.

\paragraph{Ego-relative pose} links agents to the ego vehicle through coarse relative position and orientation.
It is densely annotated for all agents and describes whether an agent is in front of, behind, left, or right of the ego and whether its direction is same, opposite, or perpendicular to the ego.

\subsection{CASCADE's Human-Annotated Dataset}

In the following, we describe how the CASCADE dataset is derived from the described scene description schema.
We first explain the annotation-clip selection and human annotation procedure, then give and overview of the CASCADE dataset and its statistics.
CASCADE's stage 1 data is available at \url{https://huggingface.co/datasets/nvidia/cascade}

\subsubsection{Data Collection and Annotation}

In principle, the CASCADE annotation concept can be applied to any dataset with ego-centric driving videos.
To produce the CASCADE dataset, we annotate driving clips from the PhysicalAI-AV dataset \cite{nvidia2025physicalai}.
In a first step, we select an initial subset 2.8k 20s scenes with \emph{complex} traffic scenarios within corpus of the 300k PhysicalAI-AV clips, before we generate CASCADE annotations for those complex scenes in a multi-stage human annotation process.

\paragraph{Candidate clip selection from PhysicalAI-AV.}
To collect a candidate pool of complex, high-reasoning scenes from the PhysicalAI-AV corpus, we first manually identify non-nominal driving clips, where non-nominal clips must meet one of the following criteria:
\begin{enumerate}
    \item The ego vehicle interacts with an agent or object.
    \item The ego vehicle maneuvers in non-standard conditions, e.g. construction zone or bad weather.
\end{enumerate}
We manually inspect around 4k clips to find 1k non-nominal scenes that match the above criteria.
With 1k positive and 2k negative clip samples, we train a logistic regression binary classifier based on the clip's Cosmos-Embed1 features \cite{nvidia2023cosmosembed1}.
We then select the highest-scoring 1.8k PhysicalAI-AV scenes within the classifier's top 5\% percentile, and combine them with the manually selected 1k non-nominal scenes to get an initial annotation pool of 2.8k complex clips.

\paragraph{Human annotation.}
Due to the complexity of CASCADE annotations, we employ a two-stage review-based human annotation process for the selected complex clips.
The process is designed to keep the amount of instructions and thus mental load manageable for annotators, to achieve a high annotation quality.
In \emph{stage 1}, the ego actions and actions of interacting actors or objects are identified.
In \emph{stage 2}, additional spatial information is added for all relevant actors and objects.
For quality assurance, each stage has two sub-steps, \emph{annotation} and \emph{review}, which are performed independently.

\emph{Stage 1} focuses on annotations for ego actions and entities (agents, traffic objects, traffic lights, environments) that directly influence the ego actions or are influenced by it.
The intention is to keep the annotator's attention on the ego actions and the entities that either cause those actions are are affected by them, without the need to consider spatial arrangements.
For each entity that is added as relevant to the ego, its actions and properties are annotated.
This stage also establishes all causal relationships in the annotation through Because-of and Influenced-by links.

\begin{figure}
    \centering
    \includegraphics[width=1\linewidth]{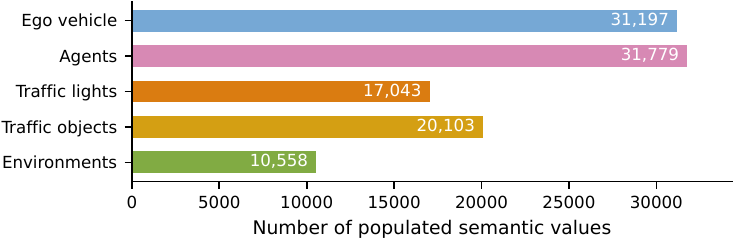}
    \caption{Populated semantic values per entity class over the 2{,}066 CASCADE-annotated clips. CASCADE annotates the dynamic actors (ego, agents) and the static scene context (traffic lights, objects, environments) at a high density.}
    \label{fig:plots_total_datapoints}
\end{figure}

\begin{figure}
    \centering
    \includegraphics[width=1\linewidth]{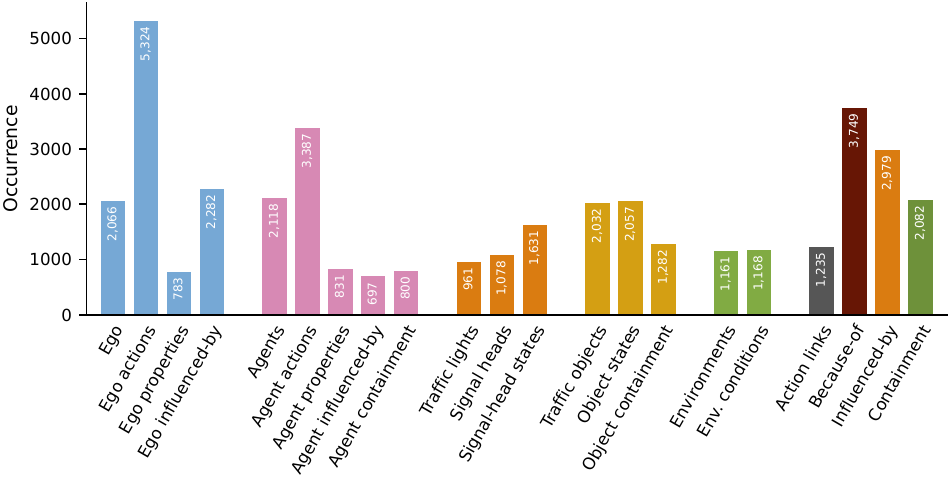}
    \caption{Populated timeline track-elements per annotation category, categories in analogy to \cref{fig:CASCADE_schema_doublecol}. Each entity class is characterized by several parallel tracks (type, actions, properties, states). Entities are connected via linkage mechanisms, counted on the right (action, because-of, influenced-by and containment). Counts are agglomerated over all five entity classes.}
    \label{fig:plots_annotation_counts}
\end{figure}

\emph{Stage 2} introduces the spatial grounding for the relevant entities from Stage 1, and shifts the annotator's attention from actions to spatial arrangement.
First, all environments are added that contain one of the Stage-1 entities and were not already included as causally-relevant in Stage 1.
With this complete collection of relevant environments, annotators are then asked to establish the full spatial linkage: This includes adding containment links for all agents, traffic objects and traffic lights, and describing the ego-relative poses for all agents.

\emph{Annotation and Review} are performed by two independent pools of workers in every stage.
Annotators generate the initial annotation per stage.
Reviewers then evaluate if an annotation quality passes or fails review based on a list of review questions.
Reviewers have the option to fix minor issues before making the pass or fail decision.
Annotations that pass the review exit their respective stage and are marked as complete, while failed clips return to the original annotator for rework based on the review comments.
Reworked clips undergo review and rework until their quality passes the review.

\begin{figure}
    \centering
    \includegraphics[width=\linewidth]{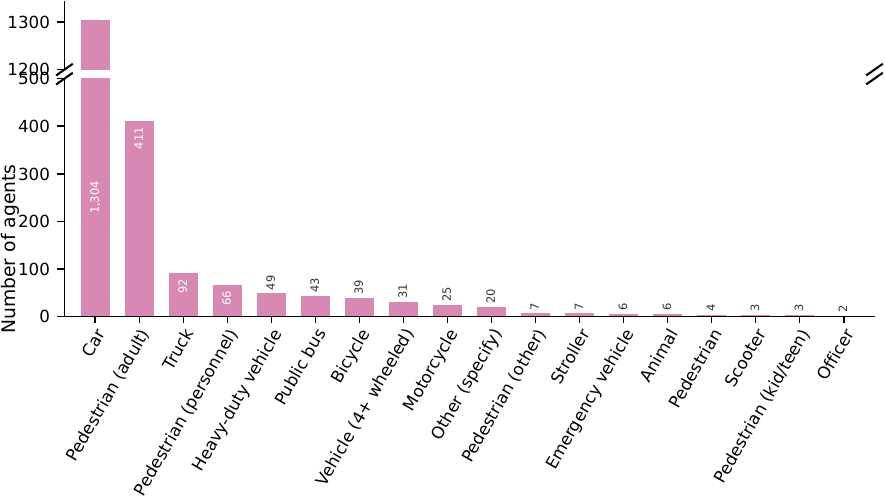}
    \caption{Distribution of the 2{,}118 annotated agents across 18 types (note the broken $y$-axis). Cars dominate, but the long tail covers vulnerable road users such as pedestrians, cyclists and scooters.}
    \label{fig:agent_types}
\end{figure}

\subsubsection{Dataset Overview}

\begin{figure*}
    \centering
    \includegraphics[height=.42\linewidth]{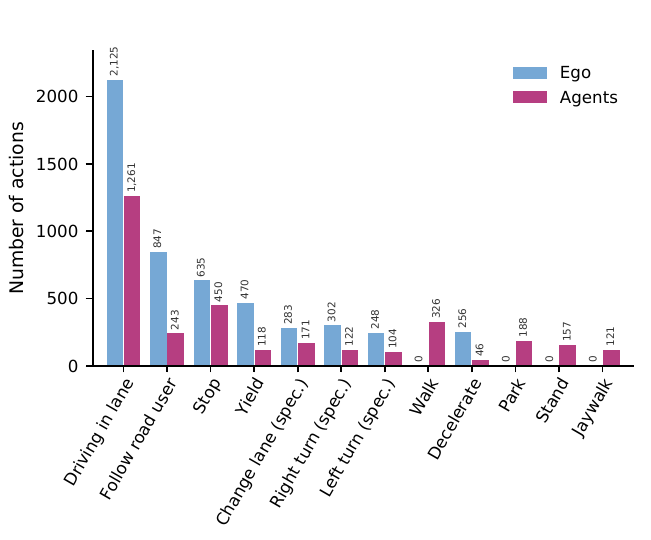}
    \hfill
    \includegraphics[height=.42\linewidth]{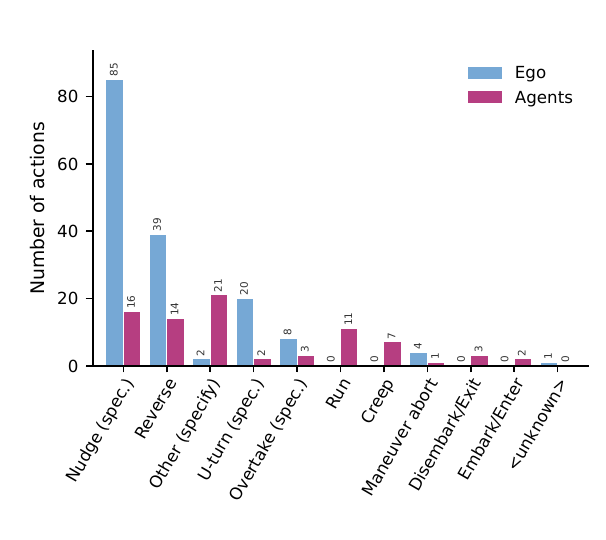}
    \vspace{-.5\baselineskip}
    \caption{Ego vs.\ agent action counts per type, split into the 12 most common (left) and the 11 rarer (right) types. Ego and agents share \emph{driving vocabulary}, while agents add \emph{pedestrian-specific actions} such as walking, parking, standing and jaywalking. For actions that come with specifications (spec.), \cref{tab:maneuver_specifications} summarizes their breakdowns into the specific sub-actions.}
    \label{fig:plots_actions}
\end{figure*}

\begin{table}[t]
  \centering
  \caption{Ego and agent action breakdowns for actions with lane, direction and safety specifications. See \cref{fig:plots_actions} for full ego and agent action statistics.}
  \label{tab:maneuver_specifications}
  \scalebox{0.8}{%
  \begin{tabular}{l@{\ \ }lrrr}
    \toprule
    \textbf{Action} & \textbf{Specification} & \textbf{Ego} & \textbf{Agent} & \textbf{Total} \\
    \midrule
    \multirow{3}{*}{\textbf{Overtake}}
      & using ego lane         & 0   & 2   & 2   \\
      & not using ego lane     & 0   & 1   & 1   \\
      & unspecified            & 8   & 0   & 8   \\
    \midrule
    \multirow{4}{*}{\textbf{Nudge}}
      & in lane                & 55  & 5   & 60  \\
      & out of lane            & 30  & 0   & 30  \\
      & out, into ego lane     & 0   & 7   & 7   \\
      & out, not into ego lane & 0   & 4   & 4   \\
    \midrule
    \multirow{2}{*}{\textbf{Change lane}}
      & left                   & 154 & 91  & 245 \\
      & right                  & 129 & 80  & 209 \\
    \midrule
    \multirow{2}{*}{\textbf{Left turn}}
      & protected              & 32  & 9   & 41  \\
      & unprotected            & 216 & 95  & 311 \\
    \midrule
    \multirow{2}{*}{\textbf{Right turn}}
      & protected              & 17  & 8   & 25  \\
      & unprotected            & 285 & 114 & 399 \\
    \midrule
    \multirow{2}{*}{\textbf{U-turn}}
      & protected              & 0   & 1   & 1   \\
      & unprotected            & 20  & 1   & 21  \\
    \bottomrule
  \end{tabular}
  }
\end{table}

\begin{figure}[t]
    \centering
    \vspace*{-1\baselineskip}
    \includegraphics[width=1\linewidth]{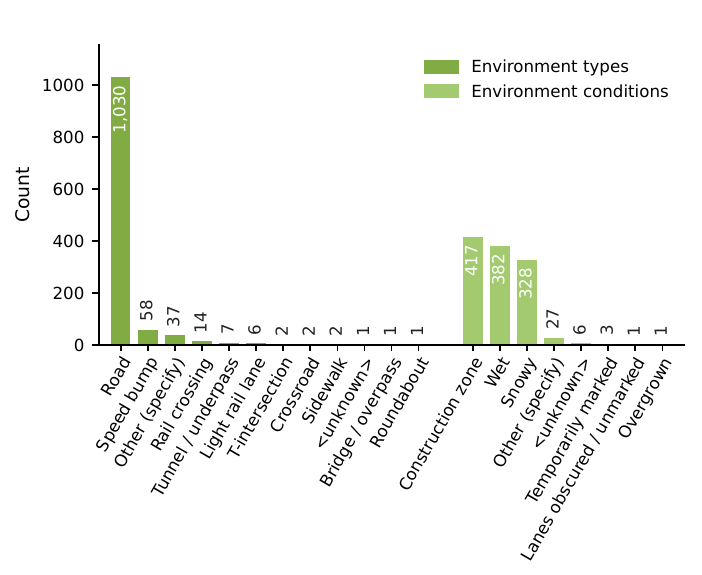}
    \vspace*{-1.5\baselineskip}
    \caption{Environment types (dark, left) and conditions (light, right). Roads dominate the types, whereas the conditions are frequently adverse, i.e.\ construction, wet or snowy.}
    \label{fig:plots_env}
\end{figure}

\begin{figure}[t]
    \centering
    \vspace*{-1\baselineskip}
    \includegraphics[width=1\linewidth]{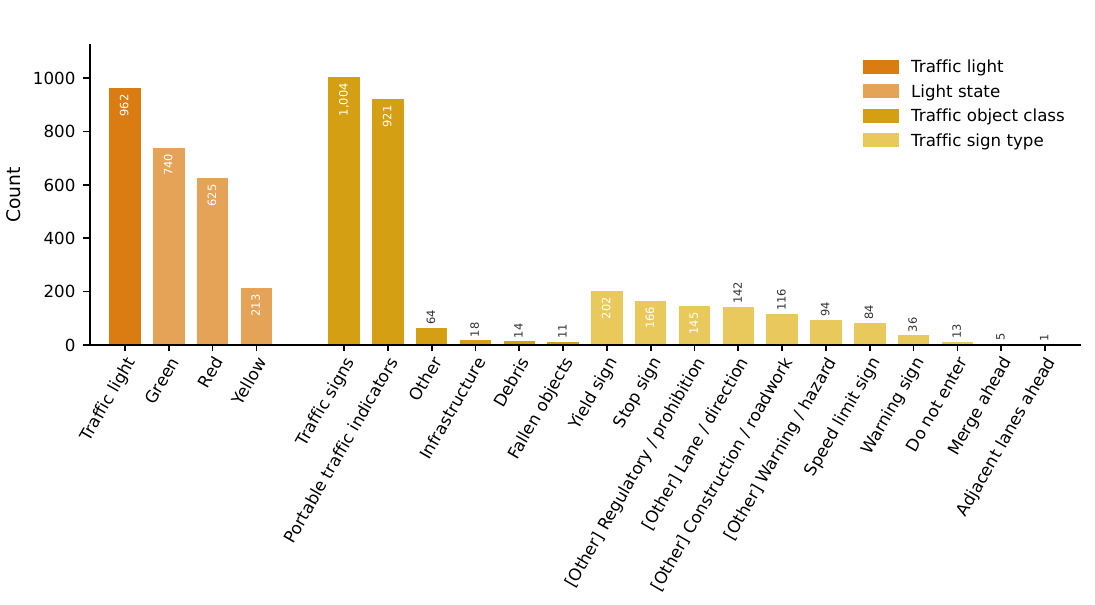}
    \vspace*{-1.5\baselineskip}
    \caption{Traffic lights and traffic objects. Left: the traffic lights and their green/red/yellow state-segment frequencies. Right: traffic objects grouped into the their coarse categories, \cf \cref{sec-supp:schema-objects}, along with detailed counts for traffic signs.}
    \label{fig:plots_tl-obj}
\end{figure}

CASCADE describes each clip as a set of rich scene semantics via actors, their actions, and the surrounding environments and objects, which are subsequently linked in space, time, and by cause.
In total, we annotate 2{,}066 train and validation clips with more than 110{,}000 populated semantic values, on average 54 per clip, distributed over the five entity classes.
Fig.~\ref{fig:plots_total_datapoints} breaks these values down per class.
The dynamic actors, ego and agents, are labeled most densely on an action level, yet the static scene context also contributes about 50\% of the annotation weight with traffic lights, traffic objects, and environment descriptions.
This illustrates that CASCADE characterizes the whole scene rather than the ego alone.\footnote{This data release provides Stage 1 annotations only; with the release of Stage 2, more annotations of ego-relative pose and environments will be introduced in Fig.~\ref{fig:plots_annotation_counts}.}

Each class is described along several parallel tracks.
Fig.~\ref{fig:plots_annotation_counts} lists the populated track-elements per category: every actor carries a type (the \texttt{Ego} or \texttt{Agent}), a dense action track with multiple actions (5{,}324 ego and 3{,}387 agent total action segments) and properties, while objects and lights additionally track time-varying states (2{,}057 object and 1{,}631 signal-head states).
These per-entity semantics are then tied together by the linkage tracks at right -- causal links with 3{,}749 because-of and 2{,}979 influenced-by connections, 1{,}235 action links that connect interdependent actions (like \texttt{yield to <what?>}), and 2{,}082 containment links.
Those realize the temporal, spatial and causal structure detailed below.
CASCADE thus records not only what populates a scene, but how its elements relate.

\subsubsection{Dataset Detailed Statistics}
Next, we analyze the annotations from scene semantics with actors and their actions, over the scene context, to the links that connect all of them.

\paragraph{Actors and their actions.}
Fig.~\ref{fig:agent_types} shows the distribution of all annotated agents over their 18 types.
Cars dominate the annotated agents, vulnerable road users like pedestrians, cyclists, motorcyclists, scooters, and strollers still make up about one fourth of the agents.
Roughly one salient agent is annotated per clip, since agents are annotated by their relevance to the ego rather than exhaustively.
\Cref{fig:plots_actions} then summarizes the action types for both the ego vehicles and all other agents.
Both groups share a driving vocabulary, \eg driving-in-lane, following, stopping, yielding and turning, but agents contribute additional pedestrian-specific actions that never apply to the ego, such as walking, standing and jaywalking.
Directional actions like turning, nudging or overtaking may come with specifications such as in lane, out of lane, protected or unprotected, which broken down in
\cref{tab:maneuver_specifications} into their fine-grained specifics.
Overall, the 8,711 action segments across ego vehicle and agents cover more than 30 types, from frequent lane-keeping down to rare actions like reversing or running.

\paragraph{Environments and Objects.}
\Cref{fig:plots_env} summarizes the annotated environment types and their conditions.
Roads dominate the environment types, but the conditions are frequently adverse: construction zones, wet and snowy surfaces constitute the majority of the 1{,}165 condition annotations, i.e.\ a large fraction of clips depict non-trivial situations.

\Cref{fig:plots_tl-obj} summarizes the statistics of traffic lights and traffic object types within the annotations.
Traffic lights (left) have dense states, showing mostly green and red, but the annotations also capture 213 yellow states.
Among traffic objects (right), traffic signs and portable indicators such as cones and barriers form the vast majority, leaving only a small remainder of infrastructure, debris and fallen objects.
We break the dominating traffic sign bucket further down by type: yield and stop signs lead, but the largest contribution are about 500 free-text ``other'' signs. For this plot, we group this free-text long tail into regulatory/prohibition, lane/direction, construction/roadwork and warning/hazard for display purposes.
This shows that annotators captured real-world signage well beyond the fixed schema categories.

\begin{table}[tb]
    \centering
    \caption{Influenced-by counts. A single influenced-by \emph{entry} (timeline row, the unit in Fig.~\ref{fig:plots_annotation_counts}) may name several influences. Of the 2{,}979 populated entries, 2{,}646 name a single influence and 333 name two to seven; the entries thus carry 3{,}451 references.}
    \label{tab:influencers_per_entry}
    \centering
    \scalebox{0.92}{
    \small
    \setlength{\tabcolsep}{4.5pt}
    \begin{tabular}{@{}lccccccc@{\hskip 8pt}c@{}}
        \toprule
        Influences/Entry              & 1       & 2   & 3   & 4  & 5  & 6  & 7 & $\Sigma$ \\
        \midrule
        Entries                            & 2{,}646 & 245 & 59  & 12 & 13 & 3  & 1 & 2{,}979 \\
        Total Influences       & 2{,}646 & 490 & 177 & 48 & 65 & 18 & 7 & 3{,}451 \\
        \bottomrule
    \end{tabular}}%
\end{table}

\begin{table}[tb]
  \caption{Action-target counts by action type and target, split by
  the actor (E ego-vehicle, A agent) that performs an action on target.
  \enquote{\emph{E} (Ego) follows \emph{Agent} in 862 annotations.}
  The \emph{Ego} can only be targeted by other agent \emph{A}, thus has only one column.}
  \label{tab:action_targets}
  \scalebox{0.79}{
  \begin{tabular}{l c r@{\ \ }r r@{\ \ \ }r r@{\ }r}
    \toprule
    & A & E & A & E & A & E & A \\
    \cmidrule(lr){2-2}\cmidrule(lr){3-4}\cmidrule(lr){5-6}\cmidrule(lr){7-8}
     Action & Ego & \multicolumn{2}{c}{Agent} & \multicolumn{2}{c}{Obj.}
      & \multicolumn{2}{c}{Unk.} \\
    \midrule
    Follow road user               & 0 & 862 & 66  & 0 & 0 & 3 & 178 \\
    Nudge (in lane)                & 0 & 56  & 5   & 6 & 0 & 0 & 0   \\
    Nudge (out of lane)            & 0 & 27  & 0   & 3 & 0 & 1 & 0   \\
    Nudge (out, into ego lane)     & 0 & 0   & 5   & 0 & 3 & 0 & 0   \\
    Nudge (out, not into ego lane) & 0 & 0   & 3   & 0 & 1 & 0 & 0   \\
    Overtake                       & 0 & 8   & 0   & 0 & 0 & 0 & 0   \\
    Overtake (using ego lane)      & 1 & 0   & 1   & 0 & 0 & 0 & 0   \\
    Overtake (not using ego lane)  & 0 & 0   & 0   & 0 & 0 & 0 & 1   \\
    Disembark/Exit                 & 0 & 0   & 3   & 0 & 0 & 0 & 0   \\
    Embark/Enter                   & 0 & 0   & 1   & 0 & 0 & 0 & 1   \\
    \midrule
    Total                          & 1 & 953 & 84  & 9 & 4 & 4 & 180 \\
    \bottomrule
  \end{tabular}
  }
\end{table}

\begin{figure}[tb]
    \centering
    \includegraphics[width=1\linewidth]{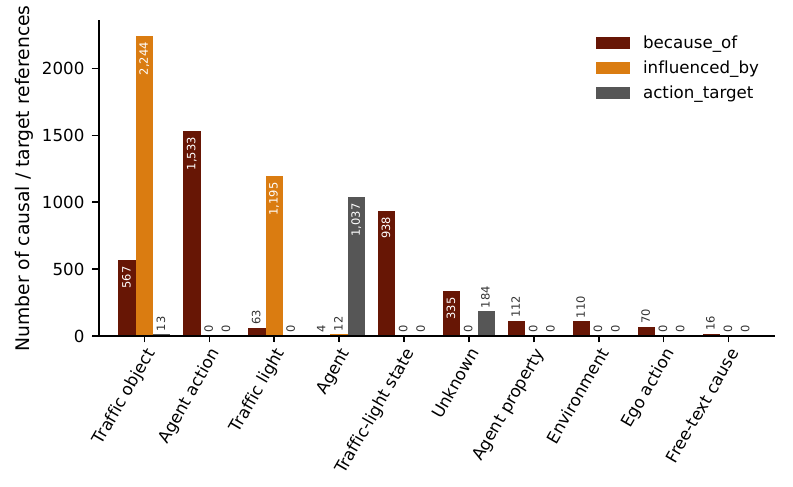}
    \caption{Number of causes (maroon), influences (orange), and action targets (gray), grouped by their connection type -- cause, influence, or action target.}
    \label{fig:plots_cause-infl}
\end{figure}

\paragraph{Causal grounding and reasoning structure.}
CASCADE connects the semantics provided by ego, agents, environments and objects through explicitly modeled causal structure.
Causally, CASCADE explicitly adds 3{,}749 because-of and 3{,}451 influenced-by references.
Each influenced-by link can carry multiple influences at once, which are individually accounted:
\Cref{tab:influencers_per_entry} shows how many influences are assigned to individual influenced-by tracks, where 2{,}979 influenced-by tracks name one influence only and 333 have multiple.
Additionally, several actions, like following, nudging and overtaking carry explicit targets (what is being followed, what is being nudged around, who is being overtaken), \cf \cref{tab:action_targets}, which adds a third causal-like layer.

\Cref{fig:plots_cause-infl} shows the dominant causes, influences and action targets across all causal links, including the soft action-target links.
The figure shows that two causal link types are complementary: influenced-by mostly points to standing infrastructure (traffic objects and traffic lights), whereas because-of points to dynamic triggers (agent actions and traffic-light states), \ie records not merely which entity but which action or signal state causes a behavior.

\begin{figure}[tb]
    \centering
    \includegraphics[width=1\linewidth]{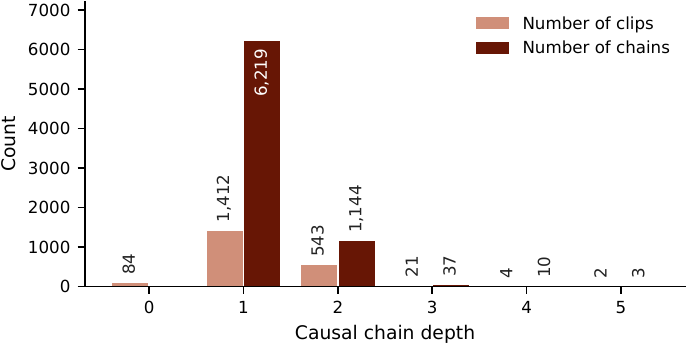}
    \caption{Causal chain depth, counted per clip (light) and per chain (dark); an edge is a because-of, influenced-by or action-target link. Most clips single-hop, with multi-hops up to depth five.}
    \label{fig:plots_cascade_depth}
\end{figure}

\begin{table}[tb]
  \centering
  \caption{Directionality of reasoning connections. Reasoning connections are causes, influences and action targets, their is cause\,$\rightarrow$\,affected. \emph{Entity} encompasses agents, traffic objects,
  traffic lights and environments; \emph{Ego} and \emph{Unknown} are treated
  separately. }
  \label{tab:link_directionality}
  \scalebox{0.8}{
  \begin{tabular}{l rrrr}
    \toprule
     \textbf{Cause\,$\rightarrow$\,Affected} & \makebox[0pt][c]{\rotatebox{30}{because\_of}}
      & \makebox[0pt][c]{\rotatebox{30}{influenced\_by}}
      & \makebox[0pt][c]{\rotatebox{30}{action\_target}} & \textbf{Total} \\
    \midrule
    Ego $\rightarrow$ Entity     & 69      & 0       & 1   & 70      \\
    \midrule
    Entity $\rightarrow$ Ego     & 2{,}742 & 2{,}655 & 962 & 6{,}359 \\
    Entity $\rightarrow$ Entity  & 585     & 796     & 88  & 1{,}469 \\
    \midrule
    Unknown $\rightarrow$ Ego    & 98      & 0       & 4   & 102     \\
    Unknown $\rightarrow$ Entity & 237     & 0       & 180 & 417     \\
    \bottomrule
  \end{tabular}
  }
\end{table}

\begin{figure}[tb]
    \centering
    \includegraphics[width=1\linewidth]{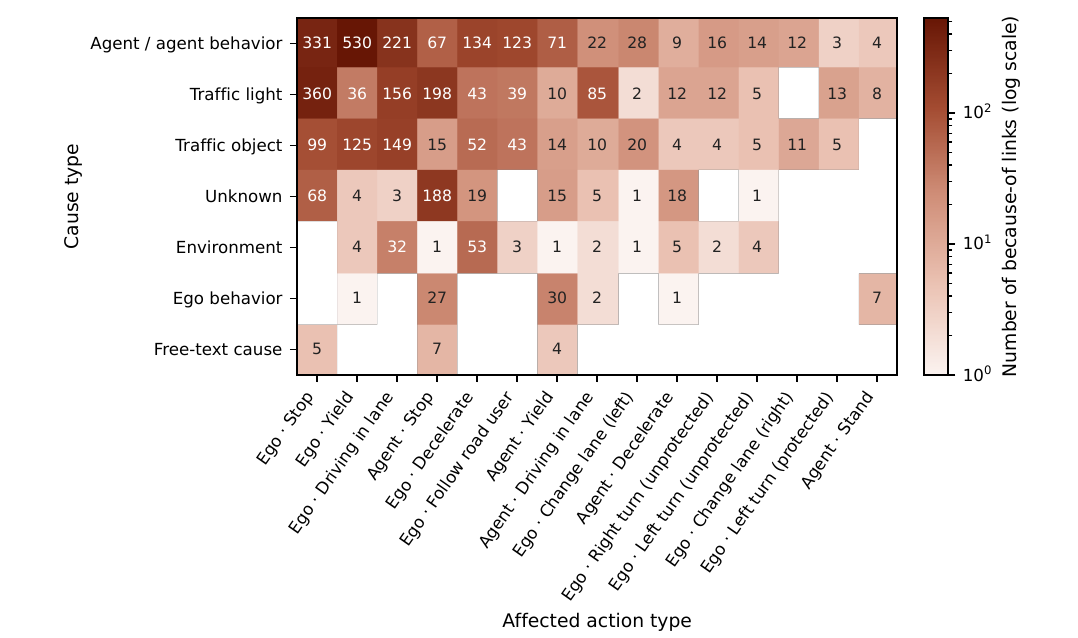}
    \caption{Most frequent causes for the most frequently affected ego- and agent actions. Note that colors are in log scale.}
    \label{fig:plots_causality}
\end{figure}

\begin{figure*}
    \centering
    \includegraphics[width=0.49\linewidth]{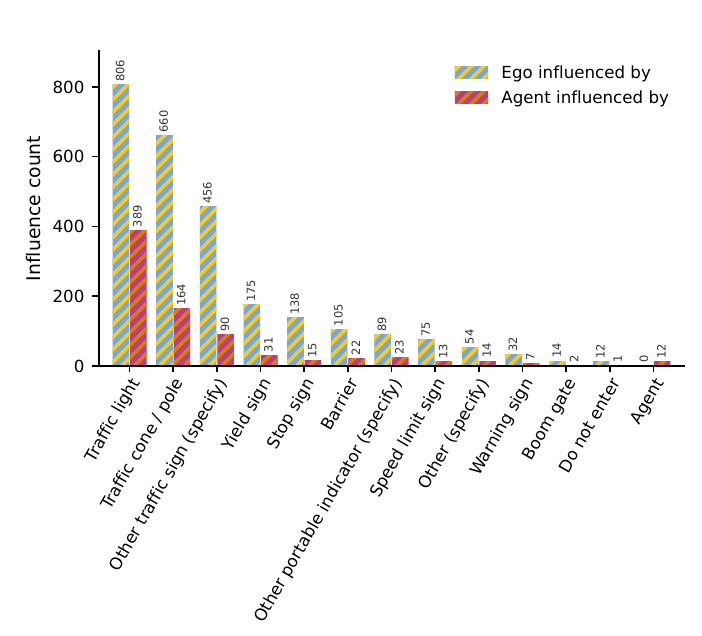}
    \includegraphics[width=.49\linewidth]{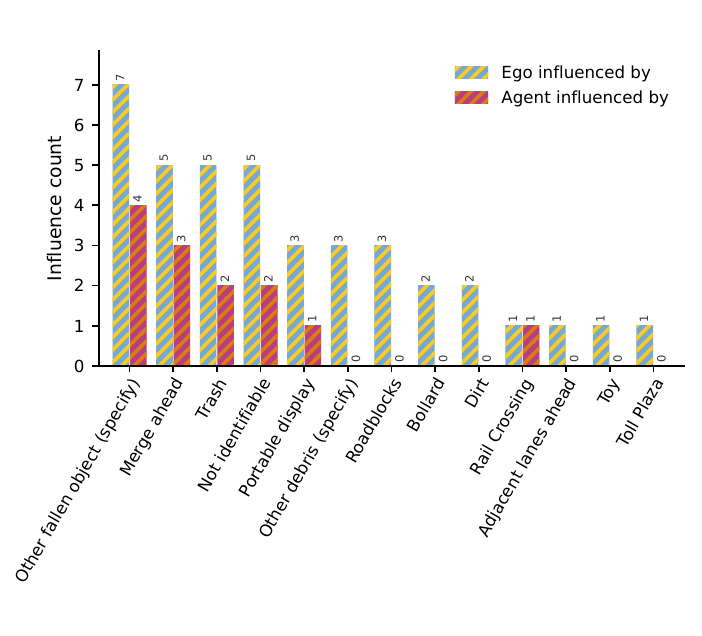}
    \vspace*{-0.75\baselineskip}
    \caption{What influences ego vs.\ agents, counted over influenced-by references and split into more common (left) and rarer (right) influence types. Regulatory infrastructure dominates the influences on ego and agents.}
    \label{fig:influences_ego-agent}
\end{figure*}

CASCADE's causal chains are established by linking agents, objects and environments that are connected through because-of, influenced-by or action-target links.
\Cref{fig:plots_cascade_depth} shows that nearly 96\% of annotated clips in the CASCADE dataset have explicit causal dependencies, and only 84 clips carry none.
While most causal chains are single hop, a substantial 28\% share of the clips have causal chains with depth 2 or more.
This clearly shows that multi-cause chains are not long-tail events but relatively frequent, and should be accounted for by reasoning evaluations.
\Cref{tab:link_directionality} analyzes the directionality of the connections in CASCADE's reasoning graphs.
Annotations are dominated by entities causing the ego to react, but the data also includes a smaller number cases where the ego affects another entity and demonstrates that bi-directional causes are worth capturing.

\Cref{fig:plots_causality} breaks down the because-of links more explicitly via a 
cause-to-effect matrix.
Overall, yielding and stopping are the most frequently caused actions -- either for ego, agents, traffic lights or objects.
CASCADE also explicitly marks causes as \emph{unknown} if the trigger cannot be inferred from the video (335 cases).

\Cref{fig:influences_ego-agent} performs a similar analysis for influenced-by links, and details which object types frequently (left) and infrequently (right) influence ego and agents.
Regulatory infrastructure like traffic lights, cones and signs are the dominant influences on both actors.
The ego has more influence annotations than other agents, as expected from ego-anchored annotations.

Taken together, these statistics show that CASCADE pairs a diverse, densely annotated set of scene elements with an explicit reasoning, and provides dense multi-hop reasoning annotations that are absent from prior human-labeled driving datasets.

\section{Limitations}

CASCADE is annotated on observational video, without interventions or controlled counterfactuals.
Its because-of and influenced-by links are therefore expert human causal attributions of positive dependencies rather than interventionally verified causation \cite{woodward2003making}, because records why an action occurred, not why an alternative did not.
Thus, CASCADE is a human-adjudicated ground truth for reasoning evaluation, without automatic causal discovery.

Further, CASCADE's causal chains are ego-rooted.
They trace causes of ego actions and the ego's effects on others, but not interactions among other actors that never involve the ego.
Annotating those requires annotating significantly more actors per scene; with roughly 1,500 hours of annotation and review, and additional 300 training hours, this was a conscious scope decision for directing human annotation to where current driving models may profit most.

\section{Conclusion}

We introduced {CASCADE}, a spatio-temporal-causal scene graph representation and human-annotated dataset for reasoning evaluation in driving.
CASCADE is motivated by the lack of structured data for assessing how well driving models reason.
Such data requires a reference for the dependencies reasoning models should recover, which must meet four requirements: spatio-temporal-causal grounding, verifiability, complete causal context, and human annotations.
CASCADE's representation addresses the first two: it grounds reasoning where, when, and why an action happened through environments and ego-relative poses, frame-by-frame time segments, and its typed reasoning graph structure makes predicted graphs verifiable against ground truth element by element, without (M)LLM judges.
CASCADE's dataset then addresses the remaining two requirements: it captures the ego's complete causal context -- causal chains of arbitrary depth in both directions, with full action and property traces for every actor along them -- and is annotated and reviewed entirely by human experts.
It instantiates the representation on 2,066 real-world clips with over 34k reasoning-relevant elements described by over 110k annotated fields.
In a field where reasoning supervision is largely auto-generated, CASCADE offers a human reference dataset to benchmark both model reasoning and pseudo-label quality, a prerequisite for reliable reasoning supervision at scale.

\subsection*{Acknowledgments}
We thank Nadine Chang, April Yang, Roberto Amoroso, Maying Shen, Philipp Hermann, Mo Hekmat, Jason Sun, Wael Elhaddad, Srinivas Venkatanarayanan, Johnny Israeli, Chintan Intwala for feedback on the annotation schema and support in the initial clip selection.
We further thank Srinivas Venkatanarayanan, Shawn Rider, Paras Kapoor, Kyle Yumen, Rajiv Bhatt, Shrilesh Shinde, Sanjeev Kumar for their support in the human annotation effort, and Naibing Du, Mario Geddes, Rochelle Woods, Laya Sleiman, Saori Kaji, Elizabeth Sanville, Ekram Mukbil, Elena Lantz, Aaraadhya Narra for helping with the business, legal and release processes.

\clearpage
\appendix
\twocolumn[{%
    {\titlefont Supplementary Material}
    \vskip20pt
}]
\setcounter{section}{0}
\setcounter{figure}{0}
\setcounter{table}{0}
\setcounter{equation}{0}
\renewcommand{\thesection}{\Alph{section}}
\renewcommand{\thefigure}{S\arabic{figure}}
\renewcommand{\thetable}{S\arabic{table}}
\renewcommand{\theequation}{S\arabic{equation}}
\newlength{\optgroupindent}\setlength{\optgroupindent}{1.5em}
\newlength{\optgrouphang}\setlength{\optgrouphang}{1.5em}
\newlength{\schemaitemsep}
\newenvironment{schemalist}[1][0pt]
  {\leavevmode\par\nopagebreak\vspace{3pt}\begingroup\setlength{\parindent}{0pt}\setlength{\parskip}{1pt}%
   \setlength{\schemaitemsep}{#1}\def\schemafirstitem{}%
   \def\item[##1]{\par\ifx\schemafirstitem\empty\def\schemafirstitem{x}\else\vspace{\schemaitemsep}\fi\leftskip=0pt\relax\textbf{\texttt{##1}}\hspace{0.5em}\ignorespaces}}
  {\par\endgroup\vspace{2pt}}
\newcommand{\opt}[1]{#1}
\newcommand{\optgroup}[1]{\par\leftskip=\optgroupindent\hangindent=\optgrouphang\hangafter=1\relax\texttt{#1:}~\ignorespaces}
\newcommand{\wire}[1]{\texttt{#1}}
\newcommand{\sitem}[1]{\textbf{\texttt{#1}}}
\newcommand{\lm}[1]{\sitem{#1}}
\newcommand{\ogroup}[1]{\texttt{#1}}

\section{CASCADE Annotation Schema}
\label{sec-supp:schema}

This appendix specifies the complete CASCADE annotation schema.
A CASCADE annotation is a reasoning scene graph (\cref{fig:CASCADE-annotation-schema}): its nodes are the \emph{entities} of the driving scene -- the ego vehicle, agents, traffic objects, traffic lights, and environments -- and its edges are the \emph{linkage mechanisms} that connect them causally and spatially.
\Cref{sec-supp:schema-entities} specifies the five entities, \cref{sec-supp:schema-linkages} the four linkage mechanisms, and \cref{sec-supp:schema-conventions} the conventions that cut across both (time segments, flags, keypoints) together with the clip-level metadata.

\subsection{Entities}
\label{sec-supp:schema-entities}

Every entity is described by the same three building blocks:
\begin{itemize}[leftmargin=1.6em, itemsep=1pt, topsep=2pt]
    \item \textbf{Type properties} -- static attributes fixed for the whole clip: the entity's \sitem{Type} and, where applicable, its \sitem{Amount}.
    \item \textbf{Annotation tracks} -- time-dependent annotations that are grounded only in time: \sitem{Actions}, \sitem{Properties}, \sitem{State}, and \sitem{Conditions}.
    \item \textbf{Linkage mechanisms} -- time-dependent links that ground an entity in other entities: \lm{Because-of}, \lm{Influenced-by}, \lm{Containment}, and \lm{Ego-relative Pose}.
\end{itemize}
\Cref{tab-supp:schema-overview} summarizes which building blocks each entity uses.
The entity descriptions below follow the same three paragraphs and list every option the annotation tool offers; options marked \opt{Other} require a free-text description.

\begin{table*}[t]
\centering\small
\scalebox{0.92}{%
\begin{tabular}{@{}lllll@{}}
\toprule
 & \multicolumn{2}{l}{\textbf{Type properties}} & \textbf{Annotation tracks} & \textbf{Linkage mechanisms} \\
\textbf{Entity} & Type & Amount & & \\
\midrule
Ego vehicle     & singleton        & --         & Actions, Properties & Because-of, Influenced-by, Containment \\
Agents          & 19 types         & 2 options  & Actions, Properties & Because-of, Influenced-by, Containment, Ego-rel.\ Pose \\
Traffic objects & 23 types         & 5 options  & State               & Containment \\
Traffic lights  & 1 (signal heads) & --         & State               & Containment (location, lane control) \\
Environments    & 24 types         & --         & Conditions          & -- \\
\bottomrule
\end{tabular}}
\caption{Building blocks per entity. Every element of an annotation track or linkage mechanism is a time segment; type properties are constant per entity. Environments are the target of Containment and Because-of links but carry none themselves.}
\label{tab-supp:schema-overview}
\end{table*}

\subsubsection{Ego Vehicle}
\label{sec-supp:schema-ego}

The dashcam vehicle. Exactly one per clip, implicitly present for the full clip duration (no visibility window).

\paragraph{Type properties.}
\begin{schemalist}
    \item[Type] \opt{Ego} (singleton).
\end{schemalist}

\paragraph{Annotation tracks.}
\begin{schemalist}
    \item[Actions] dense, one action at a time.
        \optgroup{Longitudinal} \opt{Driving in lane}, \opt{Decelerate}, \opt{Stop}, \opt{Creep}, \opt{Reverse}, \opt{Yield}, \opt{Follow road user}, \opt{Maneuver abort}.
        \optgroup{Lateral} \opt{Change lane (left)}, \opt{Change lane (right)}, \opt{Nudge (in lane)}, \opt{Nudge (out of lane)}, \opt{Overtake}.
        \optgroup{Turns} \opt{Left turn}, \opt{Right turn}, \opt{U-turn}, each as \opt{protected} or \opt{unprotected}; \opt{Other turn}.
        \optgroup{Transitions} \opt{Enter}, \opt{Exit}.
        \optgroup{Fallback} \opt{Other}.
        \optgroup{Action links} Follow road user and Overtake link to the followed/overtaken agent; Nudge, Enter, and Exit link to the agent or traffic object they relate to (or \opt{Unknown}).
    \item[Properties] sparse, may overlap actions; same vocabulary as for agents (\cref{sec-supp:schema-agents}).
        \optgroup{All} \opt{Stopped}, \opt{Slow}, \opt{Fast}, \opt{Aggressive}, \opt{Erratic}, \opt{Emergency}, \opt{On duty}, \opt{Double parked}, \opt{Outside camera view}, \opt{Signal}, \opt{Other}.
\end{schemalist}

\paragraph{Linkage mechanisms.}
\begin{schemalist}
    \item[Because-of] on every action; mandatory for \opt{Stop}, \opt{Decelerate}, and \opt{Yield}.
        \optgroup{Targets} \opt{Agent action}, \opt{Ego action}, \opt{Agent property}, \opt{Traffic-light state}, \opt{Traffic light}, \opt{Traffic object}, \opt{Environment}, \opt{Unknown}, \opt{Other}.
    \item[Influenced-by] sparse, attached to the ego as a whole.
        \optgroup{Targets} \opt{Traffic object}, \opt{Traffic light}.
    \item[Containment] dense; the ego is in at least one environment at every moment.
        \optgroup{Targets} \opt{Environment} (with lane).
    \item[Ego-relative Pose] not applicable.
\end{schemalist}

\subsubsection{Agents}
\label{sec-supp:schema-agents}

Road users that interact with or influence the ego, including parked vehicles. Each agent is annotated for its visibility window.

\paragraph{Type properties.}
\begin{schemalist}
    \item[Type] one road-user class per agent; the unspecified \opt{Vehicle}/\opt{Pedestrian} entries are used when the sub-type cannot be determined.
        \optgroup{Vehicles} \opt{Vehicle} (unspecified), \opt{Car}, \opt{Truck}, \opt{Heavy-duty vehicle}, \opt{Public bus}, \opt{Emergency vehicle}, \opt{Motorcycle}, \opt{Scooter}, \opt{Bicycle}.
        \optgroup{Pedestrians} \opt{Pedestrian} (unspecified), \opt{Adult}, \opt{Kid/Teen}, \opt{Officer}, \opt{Personnel}, \opt{Stroller}, \opt{Wheelchair}, \opt{Other}.
        \optgroup{Other} \opt{Animal} (with description), \opt{Other}.
    \item[Amount] \opt{Single}, \opt{Row/group}.
\end{schemalist}

\paragraph{Annotation tracks.}
\begin{schemalist}
    \item[Actions] dense, one action at a time; the vocabulary depends on the agent type.
        \optgroup{All agents} \opt{Decelerate}, \opt{Stop}, \opt{Creep}, \opt{Yield}, \opt{Follow road user}, \opt{Maneuver abort}, \opt{Enter}, \opt{Exit}, \opt{Other}.
        \optgroup{Vehicles only} \opt{Driving in lane}, \opt{Park}, \opt{Reverse}, \opt{Change lane (left)}, \opt{Change lane (right)}, \opt{Nudge (in lane)}, \opt{Nudge (out of lane: into ego lane)}, \opt{Nudge (out of lane: not into ego lane)}, \opt{Overtake (using ego lane)}, \opt{Overtake (not using ego lane)}, \opt{Left turn}, \opt{Right turn}, \opt{U-turn} (each \opt{protected} or \opt{unprotected}), \opt{Other turn}.
        \optgroup{Pedestrians and animals only} \opt{Stand}, \opt{Walk}, \opt{Run}, \opt{Jaywalk}.
        \optgroup{Action links} as for the ego (\cref{sec-supp:schema-ego}); links may point to other agents, the ego, traffic objects, or \opt{Unknown}.
    \item[Properties] sparse, may overlap actions. A \opt{Signal} property additionally records its source, its intent, and the addressed road users.
        \optgroup{All agents} \opt{Stopped}, \opt{Slow}, \opt{Fast}, \opt{Aggressive}, \opt{Erratic}, \opt{Emergency}, \opt{On duty}, \opt{Double parked}, \opt{Outside camera view}, \opt{Signal}, \opt{Other}.
        \optgroup{Signal source} \opt{Flashing light}, \opt{Hand gesture}, \opt{Holding sign} (sign: \opt{Stop}, \opt{Yield}, \opt{Slow}, \opt{Not identifiable}, \opt{Other}; flag if not facing ego), \opt{Other}.
        \optgroup{Signal intent} \opt{Left indicator}, \opt{Right indicator}, \opt{Stop}, \opt{Slow down}, \opt{Proceed}, \opt{Follow}, \opt{Caution}, \opt{Danger}, \opt{Unclear/incorrectly used}, \opt{Other}.
        \optgroup{Signal link} the addressed agents, the ego, or \opt{Unknown}.
\end{schemalist}

\paragraph{Linkage mechanisms.}
\begin{schemalist}
    \item[Because-of] on every action; mandatory for \opt{Stop}, \opt{Decelerate}, and \opt{Yield}.
        \optgroup{Targets} \opt{Ego action}, \opt{Agent action}, \opt{Agent property}, \opt{Traffic-light state}, \opt{Traffic light}, \opt{Traffic object}, \opt{Environment}, \opt{Unknown}, \opt{Other}.
    \item[Influenced-by] sparse, attached to the agent as a whole.
        \optgroup{Targets} \opt{Traffic object}, \opt{Traffic light}, \opt{Agent}.
    \item[Containment] dense; an agent is in at least one environment at every moment of its visibility.
        \optgroup{Targets} \opt{Environment} (with lane).
    \item[Ego-relative Pose] dense over the visibility window.
        \optgroup{Targets} \opt{Ego} (with position and direction, \cref{sec-supp:schema-linkages}).
\end{schemalist}

\subsubsection{Traffic Objects}
\label{sec-supp:schema-objects}

Static and regulatory items that influence the ego's or an agent's behavior, annotated for their visibility window.

\paragraph{Type properties.}
\begin{schemalist}
    \item[Type] one object class per object.
        \optgroup{Traffic signs} \opt{Stop sign}, \opt{Yield sign}, \opt{Speed limit sign}, \opt{Merge ahead}, \opt{Adjacent lanes ahead}, \opt{Do not enter}, \opt{Warning sign}, \opt{Other traffic sign}.
        \optgroup{Infrastructure} \opt{Bollard}, \opt{Boom gate}.
        \optgroup{Portable traffic indicators} \opt{Traffic cone}, \opt{Barrier}, \opt{Roadblock}, \opt{Portable display}, \opt{Other portable indicator}.
        \optgroup{Fallen objects} \opt{Ball}, \opt{Toy}, \opt{Other fallen object}.
        \optgroup{Debris} \opt{Dirt}, \opt{Trash}, \opt{Other debris}.
        \optgroup{Other} \opt{Not identifiable}, \opt{Other}.
    \item[Amount] for objects that occur in arrangements (bollards, cones, barriers, roadblocks, portable displays, other portable indicators, other): \opt{Single}, \opt{Line/Row}, \opt{Channelizing line}, \opt{Perimeter}, \opt{Group}.
\end{schemalist}

\paragraph{Annotation tracks.}
\begin{schemalist}
    \item[State] dense over the visibility window:
        \optgroup{motion} \opt{Static}, \opt{Moving/Rolling};
        \optgroup{opening} (openable objects such as boom gates) \opt{Open}, \opt{Closed}.
\end{schemalist}

\paragraph{Linkage mechanisms.}
\begin{schemalist}
    \item[Because-of] not applicable; objects have no actions and instead appear as \ogroup{Targets} of \lm{Because-of} and \lm{Influenced-by}.
    \item[Influenced-by] not applicable.
    \item[Containment] dense; typically with the \opt{near} flag and an \opt{edge} side for objects placed at the road edge.
        \optgroup{Targets} \opt{Environment} (with lane).
    \item[Ego-relative Pose] not applicable.
\end{schemalist}

\subsubsection{Traffic Lights}
\label{sec-supp:schema-lights}

Signalized control, annotated for the visibility window of the physical structure.

\paragraph{Type properties.}
\begin{schemalist}
    \item[Type] \opt{Traffic light}, structured as one physical light structure holding one or more \emph{signal heads} (lamp clusters); every annotation track and linkage below is recorded per signal head.
\end{schemalist}

\paragraph{Annotation tracks.}
\begin{schemalist}
    \item[State] dense per signal head:
        \optgroup{color} \opt{Red}, \opt{Yellow}, \opt{Green}, \opt{Other};
        \optgroup{shape} \opt{Round}, \opt{Arrow left}, \opt{Arrow right}, \opt{Arrow up}, \opt{Arrow down}, \opt{Other};
        \optgroup{mode} \opt{Fixed}, \opt{Flashing}, \opt{Off}.
        For \opt{Yellow} states three yes/no questions are added: is the yellow light on the ego's path, is the ego in the intersection while yellow, and could the ego have cleared safely.
\end{schemalist}

\paragraph{Linkage mechanisms.}
\begin{schemalist}
    \item[Because-of] not applicable (light states appear as \ogroup{Targets} of \lm{Because-of}).
    \item[Influenced-by] not applicable (lights appear as \ogroup{Targets} of \lm{Influenced-by}).
    \item[Containment] two flavors: the \opt{location} of the structure and, per signal head, the \opt{lane control} declaring which lanes the head governs.
        \optgroup{Targets (location)} \opt{Environment}.
        \optgroup{Targets (lane control)} \opt{Environment} (with one or more lanes).
    \item[Ego-relative Pose] not applicable.
\end{schemalist}

\subsubsection{Environments}
\label{sec-supp:schema-environments}

Road infrastructure that the ego and agents move through, annotated for as long as visible. Several environments are typically active at once (e.g.\ a road and the intersection it leads into).

\paragraph{Type properties.}
\begin{schemalist}
    \item[Type] one class per environment; each environment additionally records its number of lanes (lane merges and forks also the number of outgoing lanes) and a one-way flag.
        \optgroup{Roads} \opt{Road}, \opt{Lane merge}, \opt{Lane fork}, \opt{Bridge}, \opt{Tunnel}, \opt{Roundabout}.
        \optgroup{Intersections} \opt{T-intersection}, \opt{Y-intersection}, \opt{Crossroad (4-way)}, \opt{5-way}, \opt{6-way}, \opt{6+-way}, \opt{Other intersection}.
        \optgroup{Road sides} \opt{Paved shoulder}, \opt{Grass shoulder}, \opt{Sidewalk}, \opt{Cycle lane}.
        \optgroup{Crossings and special infrastructure} \opt{Pedestrian crossing}, \opt{Rail crossing}, \opt{Light-rail lane}, \opt{Speed bump}, \opt{Garage}, \opt{Toll plaza}.
        \optgroup{Fallback} \opt{Other}.
\end{schemalist}

\paragraph{Annotation tracks.}
\begin{schemalist}
    \item[Conditions] sparse, several may overlap:
        \optgroup{temporary} \opt{Construction zone}, \opt{Temporarily marked};
        \optgroup{surface} \opt{Wet}, \opt{Snowy}, \opt{Overgrown};
        \optgroup{markings} \opt{Shared marked center lane}, \opt{No direction divider}, \opt{Lanes obscured/unmarked};
        \optgroup{fallback} \opt{Other}.
\end{schemalist}

\paragraph{Linkage mechanisms.}
\begin{schemalist}
    \item[Because-of] not applicable (environments appear as \ogroup{Targets} of \lm{Because-of}).
    \item[Influenced-by] not applicable.
    \item[Containment] not applicable as source; environments are the \ogroup{Targets} of all \lm{Containment} links.
    \item[Ego-relative Pose] not applicable.
\end{schemalist}

\subsection{Linkage Mechanisms}
\label{sec-supp:schema-linkages}

Linkage mechanisms are the edges of the reasoning scene graph. They are time segments attached to a source entity that reference one or more target elements by id, and they split into two groups by the grounding they provide (\cref{fig:CASCADE-annotation-schema}):
\begin{itemize}[leftmargin=1.6em, itemsep=1pt, topsep=2pt]
    \item \textbf{Causal links} -- ground \emph{why} an entity behaves as it does: \lm{Because-of} and \lm{Influenced-by}.
    \item \textbf{Spatial links} -- ground \emph{where} an entity is: \lm{Containment} and \lm{Ego-relative Pose}.
\end{itemize}
Temporal grounding needs no link group of its own: every link is itself a time segment, so the \emph{when} of a relation is either inherited from the time-segmented elements it connects (e.g.\ a \lm{Because-of} link spans the action it explains) or stated explicitly by the link's own segment (e.g.\ a \lm{Containment} segment bounds the time an agent occupies a lane).

\paragraph{Causal links.}
\begin{schemalist}[6pt]
    \item[Because-of] (\wire{because\_of}) -- the causal link. Attached to an \sitem{Actions} element of the ego or an agent, it lists the elements that caused the action. Multiple causes are allowed; chaining \lm{Because-of} links across actions yields multi-hop causal chains.
        \optgroup{Targets} \opt{Ego action}, \opt{Agent action}, \opt{Agent property} (e.g.\ a signal), \opt{Traffic-light state}, \opt{Traffic light}, \opt{Traffic object}, \opt{Environment}, \opt{Unknown} (cause not visible or not identifiable), \opt{Other} (free-text cause).
    \item[Influenced-by] (\wire{influenced\_by}) -- the soft causal link. Attached to the ego or an agent as a whole (not to an action), it lists elements that shape behavior over the segment without triggering a specific action (e.g.\ a speed-limit sign while driving in lane).
        \optgroup{Targets} \opt{Traffic object}, \opt{Traffic light}, \opt{Agent} (agents only).
\end{schemalist}

\paragraph{Spatial links.}
\begin{schemalist}[6pt]
    \item[Containment] (\wire{containment}) -- the spatial link into an environment. Ego and agents must be contained at all times, traffic objects for their visibility window; traffic lights use the two reduced flavors \opt{location} and \opt{lane control} (per signal head).
        \optgroup{Targets} \opt{Environment}.
        \optgroup{Attributes} lane index (counted from the left; one or more lanes for lane control), \opt{near} flag (next to rather than inside the environment) with \opt{edge} side (\opt{left}, \opt{right}), \opt{illegal} flag.
    \item[Ego-relative Pose] (\wire{ego\_relative\_pose}) -- the spatial link from an agent to the ego, dense over the agent's visibility window.
        \optgroup{Targets} \opt{Ego}.
        \optgroup{Attributes} position: \opt{In front}, \opt{Left}, \opt{Right}, \opt{Behind}; direction: \opt{Same}, \opt{Opposite}, \opt{Perpendicular (left to right)}, \opt{Perpendicular (right to left)}.
\end{schemalist}
Two further references live inside annotation tracks rather than forming linkage mechanisms of their own: \ogroup{Action links} (\wire{action\_target}) -- required on Follow road user, Overtake, Nudge, Enter, and Exit, naming whom an action follows or overtakes, what it nudges around, enters, or exits -- and the \ogroup{Signal link} (\wire{link\_to}) naming the road users a \opt{Signal} property addresses.

\subsection{Shared Conventions and Clip Metadata}
\label{sec-supp:schema-conventions}

The following properties are not tied to a single entity or linkage mechanism but apply across the whole reasoning scene graph: how its elements are anchored in time, flagged, and localized in the image, plus the metadata recorded once per clip.
\begin{schemalist}[6pt]
    \item[Segments] Every element of an annotation track or linkage mechanism carries an \wire{id}, a \wire{start\_timestamp}, and an \wire{end\_timestamp}. \sitem{Actions}, \sitem{State}, \lm{Containment}, and \lm{Ego-relative Pose} are \emph{dense}: they cover the entity's full visibility window without gaps, with exactly one action per entity at a time. \sitem{Properties}, \sitem{Conditions}, \lm{Because-of}, and \lm{Influenced-by} are sparse and may overlap.
    \item[Illegal flag] (\wire{illegal\_flag}) Attachable to any action and any Containment segment to mark a traffic-rule violation, e.g.\ running a red light or an object placed where it should not be.
    \item[Keypoints] Sparse image-space trajectory points (timestamp, $x$, $y$) on agents, traffic objects, signal heads, and environments; at least one per entity.
    \item[Free text] Every \opt{Other} option, animal types, and unknown-cause fallbacks carry a description; entities may carry an optional nickname (\wire{name}).
    \item[Clip-level metadata] A required \wire{relevancy} answer -- \opt{Nominal} (not eventful) or the reason the clip is eventful: \opt{Ego adapts} to agents/objects, \opt{Special environment} requiring care, \opt{Agent adapts} to the ego, \opt{Other} (with description) -- a free-text \wire{brief\_description} focusing on the interactions, and a \wire{driving\_judgment} of the ego (\opt{good}, \opt{neutral}, \opt{bad}).
\end{schemalist}

{
    \small
    \bibliographystyle{ieeenat_fullname}
    \bibliography{main}
}

\end{document}